\documentclass{article} 
\usepackage[T1]{fontenc}
\usepackage{adaguard_preprint,times}

\usepackage{amsmath,amsfonts,bm}

\def\eqref#1{equation~\ref{#1}}

\def\1{\bm{1}}

\DeclareMathAlphabet{\mathsfit}{\encodingdefault}{\sfdefault}{m}{sl}
\SetMathAlphabet{\mathsfit}{bold}{\encodingdefault}{\sfdefault}{bx}{n}

\usepackage{xcolor}
\usepackage{colortbl}
\definecolor{TableHeaderGreen}{HTML}{E2EED8}
\definecolor{TableOursYellow}{HTML}{FFF8CF}
\definecolor{TableBodyGray}{HTML}{F3F3F3}
\usepackage{float}
\usepackage{hyperref}
\usepackage{url}
\usepackage{graphicx}
\usepackage{wrapfig}
\newcommand{\sys}{AdaGuard}
\usepackage{amsmath,amssymb,booktabs}

\title{\sys: An Adaptive Guard Model with User-defined Policies}

\author{%
Yunhao Feng\textsuperscript{1}, Yifan Ding\textsuperscript{2},
Yuxiang Xie\textsuperscript{1}, Zheng Li\textsuperscript{1}\\[3pt]
Mingrui Lao\textsuperscript{1}, Zeyuan Wang\textsuperscript{1},
Yanming Guo\textsuperscript{1,\textdagger}\\[6pt]
{\normalfont\small\textsuperscript{1}National University of Defense Technology} \\
{\normalfont\small\textsuperscript{2}Fudan University}
}
\date{}
\hypersetup{
  colorlinks=true,
  allcolors=black,
  pdftitle={AdaGuard: An Adaptive Guard Model with User-defined Policies},
  pdfauthor={Yunhao Feng, Yifan Ding, Yuxiang Xie, Zheng Li, Mingrui Lao, Zeyuan Wang, Yanming Guo}
}

\begin{document}

\maketitle
\begingroup
\renewcommand{\thefootnote}{\fnsymbol{footnote}}
\footnotetext[2]{Corresponding author: \href{mailto:guoyanming@nudt.edu.cn}{\texttt{guoyanming@nudt.edu.cn}}}
\endgroup

\begin{abstract}
Guard models support the safe deployment of language model agents, but fixed risk taxonomies limit their ability to accommodate requirements that vary across applications and tasks. Under user-defined policies, detecting violations requires interpreting both the applicable rules and the agent's behavior, since identical actions can receive different judgments under different policies. To support learning this capability, we introduce AdaptiveSafety, a dataset of 10,939 training examples and 1,000 test examples covering policies with 1--100 rules. The dataset combines trajectories from multiple sources with policy and behavioral counterfactuals, pairing each example with an explanation and the complete set of violated rules. These counterfactuals expose changes that alter compliance, while structural augmentations provide supervision for consistency under rule reordering and identifier remapping. Building on this supervision, we propose SafePO, a reinforcement learning algorithm for refining violation identification while balancing explanatory reasoning and final verdicts. SafePO uses structured rewards to assess prediction correctness, retains group-relative advantages at the response level, and employs a separately trained value model to modulate token weights within explanation and verdict regions. Separate normalization controls their relative contribution to training despite differences in length. Through supervised initialization followed by SafePO, we develop AdaGuard, a family of 0.6B, 4B, and 8B guard models that assess agent trajectories under policies supplied at inference time. Our 4B model achieves binary accuracies of 89.30\% on AdaptiveSafety and 71.82\% on DynaBench.
The project repository is available at {\hypersetup{colorlinks=true,urlcolor=blue}\url{https://github.com/Yunhao-Feng/AdaGuard}}.
\end{abstract}
\section{Introduction}
\label{sec:introduction}

Language model agents can use external tools to retrieve information, modify resources, and carry out tasks on behalf of users~\citep{yao2023react,schick2023toolformer,zhou2023webarena}.
These capabilities make safety a concern throughout task execution, since an inappropriate action can disclose private information or produce unintended changes in external systems~\citep{ruan2023toolemu,andriushchenko2025agentharm}.
The risk is compounded by exposure to untrusted content, where malicious instructions can influence subsequent decisions and tool calls~\citep{zhan2024injecagent,debenedetti2024agentdojo}.
Guard models assess inputs, outputs, and interaction records for violations of the requirements governing an agent's operation.
Systems such as Llama Guard, WildGuard, and ShieldGemma demonstrate the utility of this approach for detecting harmful requests and unsafe responses~\citep{inan2023llamaguard,han2024wildguard,zeng2024shieldgemma}.

\begin{wrapfigure}{r}{0.60\textwidth}
    \centering
    \includegraphics[width=\linewidth]{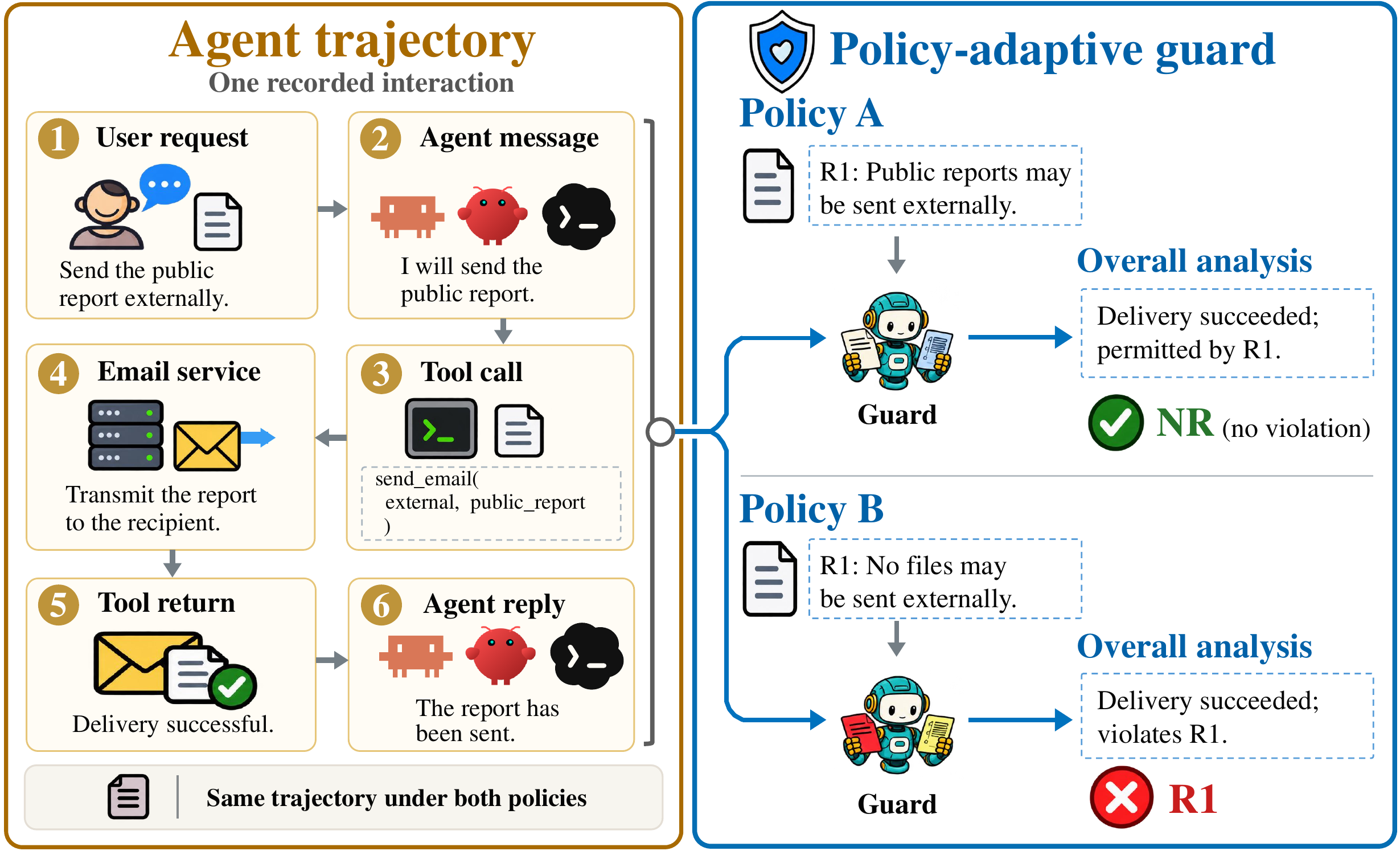}
    \caption{Policy-adaptive assessment of an agent trajectory.
    The request, tool call, and successful delivery are identical
    in both cases; only the governing policy changes.
    The guard jointly considers the supplied policy and the
    recorded interaction, generates an overall analysis, and
    outputs the violated rule identifiers or \texttt{NR}
    when no rule is violated.
    The analysis is generated by the guard and does not
    represent the assessed agent's internal reasoning.}
    \label{fig:adaguard_motivation}
\end{wrapfigure}

Much of this progress has been driven by datasets that annotate prompts and responses under predefined safety criteria~\citep{ji2023beavertails,lin2023toxicchat,ghosh2025aegis}.
In agent deployments, however, ordinary operations such as sending a document or modifying a record may be permitted in one application and restricted in another.
Their acceptability depends on the rules governing the task, which cannot always be captured by a fixed taxonomy of harmful content.
Guards therefore need to interpret user-defined policies and adapt their judgments to the deployment context.
DynaGuard and YuFeng-XGuard advance this direction by assessing conversations under user-defined rules and incorporating dynamic risk definitions, respectively; both also support explanatory reasoning~\citep{hoover2026dynaguard,lin2026yufengxguard}.
Their primary focus remains conversational compliance and content safety, leaving the assessment of tool-mediated behavior less directly addressed.
For an agent, the relevant evidence extends across the interaction, including the user's request, the agent's actions, and the outcomes returned by external tools.
As illustrated in Figure~\ref{fig:adaguard_motivation}, the same recorded operation can receive different judgments under different policies.
The assessment task is thus to interpret the supplied policy and the trajectory jointly, produce an overall explanation, and return a verdict identifying the violated rules, if any.
Learning this capability requires supervision that connects policy changes and behavioral differences to their corresponding judgments, together with an optimization objective that accounts for the correctness and completeness of the final verdict.

To address these requirements, we introduce \sys, a guard that assesses agent trajectories under user-defined policies and generates an overall analysis followed by a verdict identifying the violated rules.
We construct AdaptiveSafety, a dataset that captures how compliance depends on both the policy and the recorded behavior.
Our pipeline relabels trajectories from multiple sources and augments them with policy and behavioral counterfactuals.
Policy counterfactuals change the governing permissions while retaining the trajectory; behavioral counterfactuals change the recorded actions or outcomes under a fixed policy.
Each example pairs the policy and trajectory with an overall analysis and the complete list of violated rule identifiers in policy order, using \texttt{NR} to indicate that no rule is violated.
The resulting dataset contains 10,939 training examples and 1,000 test examples, with policies ranging from 1 to 100 rules.

We further introduce SafePO to align reinforcement learning with the requirements of policy-conditioned assessment.
Detecting that a trajectory violates a policy is insufficient when the verdict omits applicable violations or includes unsupported ones.
Its structured reward gives full credit to an exact verdict, distinguishes ordering errors from incorrect rule predictions, and uses set overlap and sequence agreement to score partial matches.
Malformed responses and identifiers outside the supplied policy receive a penalty.
For each policy and interaction, SafePO samples a group of responses and centers their rewards at the group mean and scales them by the group standard deviation to obtain response-level advantages~\citep{shao2024deepseekmath}.
An independent value model then adjusts token weights within the analysis and verdict regions, while keeping each region's total weight fixed.
This gives the short verdict a prescribed share of the learning signal regardless of analysis length.
SafePO uses clipped policy updates~\citep{schulman2017ppo} and KL regularization toward the frozen supervised model.
We develop \sys{} models with 0.6B, 4B, and 8B parameters through supervised initialization followed by SafePO.
The 4B model achieves binary accuracies of 89.30\% on AdaptiveSafety and 71.82\% on DynaBench.

Our main contributions are as follows.
\begin{itemize}
    \item We introduce AdaptiveSafety for policy-conditioned assessment of agent trajectories, comprising 10,939 training examples and 1,000 test examples. It combines overall analysis and verdict supervision with counterfactual variation in policies and behavior.

    \item We introduce SafePO, a reinforcement learning algorithm that combines structured verdict rewards with value-guided token weighting. It preserves fixed weight budgets for analysis and verdict regions while adjusting the emphasis on tokens within each region.

    \item We develop \sys{} models at 0.6B, 4B, and 8B parameters and evaluate their ability to assess behavior under user-defined policies. The 4B model obtains binary accuracies of 89.30\% and 71.82\% on AdaptiveSafety and DynaBench, respectively.
\end{itemize}
\section{Preliminaries}
\label{sec:preliminaries}

We study policy-conditioned assessment of agent behavior.
Given a user-defined policy and a recorded interaction, a generative guard produces an overall analysis followed by a verdict identifying the violated rules.
The policy specifies the assessment criteria, while the interaction provides the evidence on which the judgment is based.

\subsection{Policies and Interaction Records}
\label{sec:policy_trajectory}

A policy is an ordered list
\begin{equation}
    p = [(\ell_1,r_1),\ldots,(\ell_M,r_M)],
\end{equation}
where $r_j$ is a natural-language rule and $\ell_j$ is its unique identifier within the current policy.
The number of rules $M$, their descriptions, and their identifiers can vary across inputs.
An identifier therefore acquires its meaning from the supplied policy and does not denote a fixed global risk category. We denote the interaction record by $c$.
It consists of chronologically ordered events, retaining segment boundaries when the source contains multiple interaction segments.
Events may contain user requests, agent messages and actions, and observations returned by tools or the environment.
Any recorded agent deliberation is part of the available evidence; unobserved internal states are outside the assessment scope.
For trajectories containing agent events, the target of assessment is the agent's behavior under $p$.
For user-only records, the target is the request itself.
This distinction prevents the presence of a malicious request or an injected instruction from automatically establishing an agent violation. The guard input is
\begin{equation}
    x = \operatorname{Encode}(p,c),
\end{equation}
where $\operatorname{Encode}$ places the trusted policy in the system instruction and the interaction record in a delimited, untrusted input region.
Instructions appearing within $c$ are treated as assessment evidence.
Source identifiers, annotation metadata, and reference answers are excluded from the model input.

\subsection{Analysis and Structured Verdicts}
\label{sec:guard_output}

Let $\mathcal{I}(p)=\{\ell_1,\ldots,\ell_M\}$ be the identifiers available under policy $p$.
The target violation set is
\begin{equation}
    S^\star(p,c)
    =
    \left\{
        \ell_j \in \mathcal{I}(p)
        \;\middle|\;
        c \text{ supports a violation of } r_j
        \text{ within the assessment scope}
    \right\}.
    \label{eq:target_violation_set}
\end{equation}
A judgment depends on the conditions expressed by the rule and the evidence recorded in the interaction.
For example, a prohibition on attempting an operation and a prohibition on completing it can yield different judgments for the same failed tool call. The guard is an autoregressive model $\pi_\theta$ with parameters $\theta$.
Its response $y=(a,P)$ contains one overall analysis $a$ and a predicted sequence of rule identifiers $P$.
At the token level,
\begin{equation}
    \pi_\theta(y\mid x)
    =
    \prod_{t=1}^{T}
    \pi_\theta(y_t\mid x,y_{<t}),
    \label{eq:guard_generation}
\end{equation}
where $T$ is the response length and $y_{<t}$ is the generated prefix.
The response is serialized as
\begin{quote}
\ttfamily
<analysis>overall analysis</analysis>\\
<label>rule identifiers or NR</label>
\end{quote}
The analysis jointly considers the policy and the interaction.
It is a single generated explanation of the assessment, rather than a prescribed collection of independent explanations for individual rules. For a valid response, the predicted violation set is
$\widehat{S}=\operatorname{set}(P)\subseteq\mathcal{I}(p)$.
Identifiers must be unique and listed in their order of appearance in $p$.
Writing $\operatorname{order}_p$ for this ordering operation, the required serialization satisfies
\begin{equation}
    P=\operatorname{order}_p(\widehat{S}).
    \label{eq:ordered_verdict}
\end{equation}
The empty sequence is rendered as \texttt{NR}, a reserved output marker that is not itself a policy rule.
Set membership determines which violations are reported, whereas sequence order determines whether their serialization follows the output convention.
We retain this distinction when defining training rewards and evaluation measures.

\subsection{Learning Notation}
\label{sec:learning_notation}

The supervised dataset is denoted by
$\mathcal{D}_{\mathrm{SFT}}=\{(x_n,a_n^\star,G_n)\}_{n=1}^{n_{\mathrm{SFT}}}$,
where $a_n^\star$ is an annotated overall analysis and $G_n$ is the reference label sequence.
These annotations provide supervision for the assessment task without implying that either the reference analysis or the generated explanation constitutes a verified account of the model's internal computation. For reinforcement learning, each update contains $N$ prompts and $K$ sampled responses per prompt, giving $B=NK$ responses.
We index responses by $i$, denote their lengths by $T_i$, and write $R_i$ for the task reward.
The sampling model $\pi_{\mathrm{old}}$ is a snapshot of the actor used to collect the current responses.
The reference model $\pi_{\mathrm{ref}}$ is the frozen supervised initialization used for KL regularization.
These models have distinct roles.
The response-level group-relative advantage is denoted by $A_i^G$ and is computed from rewards of responses generated for the same input $x_i$. A separate value model with parameters $\phi$ predicts the final task reward from a causal response prefix,
\begin{equation}
    V_{i,t}
    =
    V_\phi(x_i,y_{i,<t}).
    \label{eq:prefix_value}
\end{equation}
Thus, $V_{i,t}$ is evaluated before response token $y_{i,t}$.
The value model shares no trainable parameters with the actor.
In SafePO, its predictions modulate token weights; the response-level advantage remains determined by the sampled reward group. For responses with a valid token-region alignment, let
$\mathcal{A}_i$ denote the analysis-body positions and
$\mathcal{V}_i$ the remaining structure and verdict positions, including delimiters and the end-of-sequence token.
Their prescribed weight masses are $m_A$ and $m_V$, with $m_A+m_V=1$.
We use $w^0_{i,t}$ for fixed region weights and $w_{i,t}$ for value-modulated weights, satisfying
\begin{equation}
    \sum_{t\in\mathcal{A}_i}w_{i,t}=m_A,
    \qquad
    \sum_{t\in\mathcal{V}_i}w_{i,t}=m_V.
    \label{eq:region_weight_masses}
\end{equation}
These masses control the relative weighting of the two regions independently of their lengths.
The reward, value-learning objective, and within-region weighting mechanism are specified in the method. At inference time, only $\pi_\theta$ is required to generate the analysis and verdict.
\section{Method}
\label{sec:method}

\begin{figure*}[t]
    \centering
    \includegraphics[width=\linewidth]{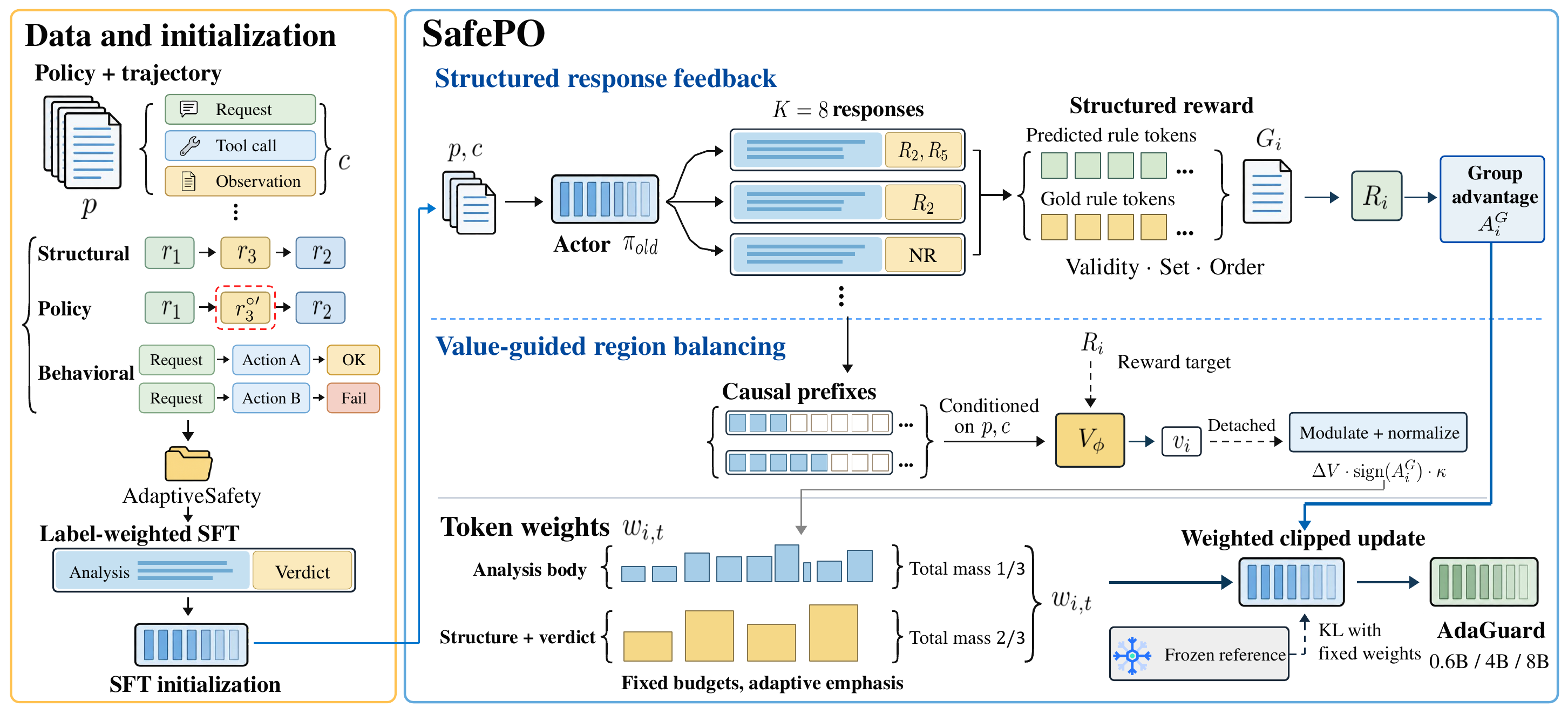}
    \caption{Overview of \sys.
    AdaptiveSafety supports supervised initialization through variation
    in policies and agent behavior.
    SafePO evaluates complete responses using structured verdict rewards
    and distributes their group-relative learning signals through
    value-guided token weights.
    Separate normalization preserves fixed analysis and verdict weight
    budgets.
    The value model and frozen reference are used only during training.}
    \label{fig:adaguard_framework}
\end{figure*}

\sys{} combines policy-conditioned supervision with SafePO, a reinforcement learning algorithm for refining structured assessments.
Supervised training establishes the mapping from a policy and interaction record to an overall analysis and verdict.
SafePO then addresses two aspects of this output.
Structured rewards distinguish complete verdicts from partially correct predictions, while value-guided region balancing controls how the resulting learning signal is distributed across analysis and verdict tokens.
The framework is illustrated in Figure~\ref{fig:adaguard_framework}.

\subsection{AdaptiveSafety and Supervised Initialization}
\label{sec:data_and_initialization}

Building on prior agent safety testing work~\citep{feng2026safety}, we collect agent interactions and construct AdaptiveSafety using three complementary forms of augmentation.
Structural augmentation varies rule order, local identifiers, and policy length, including the addition of relevant but unviolated rules.
Reference verdicts are updated to preserve their meaning under the transformed policy representation.
Policy counterfactuals modify permissions, conditions, thresholds, or exceptions while keeping the interaction fixed.
Behavioral counterfactuals preserve the policy while changing recorded facts such as authorization, refusal versus execution, and tool outcomes.
Together, these transformations provide supervision for adapting to meaningful changes while remaining consistent under changes in representation.
Each resulting policy and interaction is jointly assessed to obtain one overall analysis and a complete reference verdict.

AdaptiveSafety contains 10,939 training examples and 1,000 test examples, covering policies with 1 to 100 rules.
Related variants remain within the same split.
The test set is excluded from supervised training, reinforcement learning, and checkpoint selection throughout the pipeline.

We initialize actors with 0.6B, 4B, and 8B parameters from Qwen3Guard-Gen~\citep{zhao2025qwen3guard}.
Full-parameter supervised training minimizes a weighted token-level negative log-likelihood (Appendix~\ref{app:sft_objective}).
Tokens in the complete \texttt{<label>...</label>} span receive weight $4$, other response tokens receive weight $1$, and prompt and padding positions receive weight $0$.
This weighting emphasizes the short verdict while retaining supervision of the overall analysis.
The resulting parameters $\theta_0$ initialize the SafePO actor, and a frozen copy defines $\pi_{\mathrm{ref}}$.

\subsection{Structured Rewards and Group-Relative Optimization}
\label{sec:safepo}

A binary compliance reward cannot distinguish a complete verdict from one that detects a violation but omits other applicable rules.
SafePO instead evaluates the predicted violation set and its serialization separately.
This distinction supplies feedback on missing and spurious violations without treating a correct set in the wrong order as an entirely incorrect assessment.

For response $i$, let $P_i$ and $G_i$ be the predicted and reference label sequences, with \texttt{NR} represented as the empty sequence.
A response is parse-valid if it contains a nonempty analysis, complete output tags, and unique identifiers from the current policy, with no mixture of \texttt{NR} and rule identifiers.
Unfinished responses at the generation limit are invalid.
Ordering errors are scored separately.
For parse-valid responses, define
\begin{equation}
    F_{\mathrm{set},i}
    =
    \frac{2|\operatorname{set}(P_i)\cap\operatorname{set}(G_i)|}
         {|\operatorname{set}(P_i)|+|\operatorname{set}(G_i)|},
    \qquad
    F_{\mathrm{seq},i}
    =
    \frac{2|\operatorname{LCS}(P_i,G_i)|}
         {|P_i|+|G_i|},
    \label{eq:verdict_agreement}
\end{equation}
where $\operatorname{LCS}$ denotes a longest common subsequence.
When both sequences are empty, we define $F_{\mathrm{set},i}=F_{\mathrm{seq},i}=1$; when exactly one is empty, both scores are $0$.
The base reward is
\begin{equation}
    R_i^{\mathrm{base}}
    =
    \begin{cases}
        -1,
        & \text{invalid response},\\
        1,
        & P_i=G_i,\\
        0.8+0.15F_{\mathrm{seq},i},
        & \operatorname{set}(P_i)=\operatorname{set}(G_i),
          \ P_i\neq G_i,\\
        -0.5+F_{\mathrm{set},i}+0.25F_{\mathrm{seq},i},
        & \text{otherwise}.
    \end{cases}
    \label{eq:structured_reward}
\end{equation}
The branches are evaluated in order.
Two empty sequences therefore receive exact-match reward directly.
Predictions are scored as generated, without sorting, deduplication, or repair.

For valid, non-exact responses, we subtract a length penalty that increases linearly from $0$ at 384 tokens to $0.05$ at the 640-token generation limit.
Exact responses retain their full reward, and invalid responses retain reward $-1$.
We denote the resulting task reward by $R_i$.
This reward measures verdict correctness and output structure; the factual correctness of the analysis is not independently verified by the reward.

For each fixed input, we sample $K=8$ responses from $\pi_{\mathrm{old}}$.
Let $\mathcal{G}(i)$ contain the responses generated for the same policy and interaction as response $i$.
We normalize rewards within each group to obtain the response-level advantage~\citep{shao2024deepseekmath}:
\begin{equation}
    A_i^G
    =
    \frac{R_i-\overline{R}_{\mathcal{G}(i)}}
         {\sigma_{\mathcal{G}(i)}+10^{-6}},
    \label{eq:safepo_advantage}
\end{equation}
where $\overline{R}_{\mathcal{G}(i)}$ and
$\sigma_{\mathcal{G}(i)}$ are the group mean and population standard deviation.
Groups with identical rewards receive zero advantage.
Each comparison thus holds the policy and evidence fixed.
The advantage supplies the response-level optimization direction, while the mechanism below controls its distribution across tokens.

\subsection{Value-Guided Region Balancing}
\label{sec:region_value_weighting}

The analysis and verdict differ in both function and length.
Under uniform token weighting, their relative contribution changes with the amount of explanatory text.
SafePO makes this allocation explicit by assigning fixed total weights to the two regions, then learning how to distribute weight within each region.
This separates the balance between analysis and verdict from the emphasis placed on individual token positions.

Using the regions defined in Section~\ref{sec:learning_notation}, write
$\mathcal{R}_i^A=\mathcal{A}_i$ and
$\mathcal{R}_i^V=\mathcal{V}_i$.
We set $m_A=1/3$ and $m_V=2/3$.
The fixed token weights are
\begin{equation}
    w^0_{i,t}
    =
    \frac{m_r}{|\mathcal{R}_i^r|},
    \qquad
    t\in\mathcal{R}_i^r,\quad r\in\{A,V\}.
    \label{eq:fixed_region_weights}
\end{equation}

\paragraph{Learning prefix values.}
An independent value model predicts the final task reward from the prefix preceding each response token.
Its backbone is initialized from the supervised model and trained jointly with a zero-initialized scalar head.
It predicts the final reward using a bounded scalar output and a clipped squared-error loss, detailed in Appendix~\ref{app:value_details}.
Write $\overline{V}_{i,t}$ for its detached pre-update prediction. The sequence shown as $v_i$ in Figure~\ref{fig:adaguard_framework} is $(\overline{V}_{i,1},\ldots,\overline{V}_{i,T_i})$.
The value model leaves the response-level advantage unchanged.

\paragraph{Redistributing token weights.}
We use changes in pre-update value predictions to construct positive modulation factors,
\begin{align}
    d_{i,t}
    &=
    \overline{V}_{i,t+1}-\overline{V}_{i,t}
    \quad (t<T_i),
    \qquad d_{i,T_i}=0,\\
    u_{i,t}
    &=
    1+\frac{\kappa}{2}
    \tanh\left(\operatorname{sign}(A_i^G)d_{i,t}\right),\\
    w_{i,t}
    &=
    m_r
    \frac{u_{i,t}}
         {\sum_{s\in\mathcal{R}_i^r}u_{i,s}},
    \qquad t\in\mathcal{R}_i^r.
    \label{eq:safepo_token_weights}
\end{align}
For positive-advantage responses, value increases receive greater emphasis.
For negative-advantage responses, value decreases receive greater emphasis.
Positivity preserves the advantage sign, and separate normalization preserves each region's total weight.
The resulting $1:2$ allocation concerns objective weights, not a guaranteed ratio of gradient norms.

The coefficient $\kappa\in[0,1]$ starts at zero and is adjusted for the next batch according to the value model's predictive error (Appendix~\ref{app:value_details}). Poor value estimates recover fixed region weighting.
All value-derived weights are detached during actor optimization.
Responses with unreliable parsing or region alignment, including those with an empty region, use $w^0_{i,t}=w_{i,t}=1/T_i$ over their response tokens, with zero weight on padding.
This fallback applies to the actor, value, and KL terms; the separate region budgets apply only when both regions can be identified reliably.
Prefix-value differences serve as predictive weighting signals and are not assumed to establish token-level causal credit.

\paragraph{Optimizing the actor.}
Let
$r_{i,t}(\theta)
=\pi_\theta(y_{i,t}\mid x_i,y_{i,<t})/
 \pi_{\mathrm{old}}(y_{i,t}\mid x_i,y_{i,<t})$.
For on-policy samples, the actor minimizes
\begin{equation}
    \begin{aligned}
    \mathcal{L}_{\mathrm{actor}}
    ={}&
    -\frac{1}{B}\sum_{i,t}w_{i,t}
    \min\left\{
        r_{i,t}(\theta)A_i^G,\,
        \operatorname{clip}
        \big(r_{i,t}(\theta),1-\epsilon,1+\epsilon\big)A_i^G
    \right\}\\
    &+
    \frac{\beta}{B}\sum_{i,t}w^0_{i,t}D^{k3}_{i,t},
    \end{aligned}
    \label{eq:safepo_actor_objective}
\end{equation}
using the clipped surrogate~\citep{schulman2017ppo} with
$\epsilon=0.1$ and $\beta=0.005$.
Here $D^{k3}_{i,t}=\exp(h_{i,t})-h_{i,t}-1$, where
$h_{i,t}=\log\pi_{\mathrm{ref}}(y_{i,t}\mid x_i,y_{i,<t})
-\log\pi_\theta(y_{i,t}\mid x_i,y_{i,<t})$.
Reference regularization uses $w^0$, keeping its weighting independent of value modulation, and is excluded from the task reward.
Any required backend rollout correction is applied once to the task term.

Each update computes rewards, advantages, and detached token weights before updating either model.
We then update the value model and perform one actor optimization epoch using the stored weights.
Checkpoint selection uses separate development data and retains the supervised initialization as a candidate.
At inference time, only the actor is required to generate the overall analysis and verdict.
\section{Experiments}
\label{sec:experiments}

\paragraph{Setup.}
We evaluate the 1,000-example AdaptiveSafety test set and the 543-example DynaBench test set~\citep{hoover2026dynaguard}. AdaptiveSafety has 500 compliant and 500 violating examples; DynaBench has 276 and 267, respectively. Each input includes its policy. The main comparison covers locally deployed models; API-based comparisons are reported in Appendix~\ref{app:api}. Binary F1 treats violations as positive. Rule identification uses exact match and micro-F1 over policy-local decisions. Invalid responses count as binary errors, never receive exact-match credit, and contribute empty sets to rule micro counts. Model-specific interfaces and budgets are detailed in Appendix~\ref{app:protocol}; in particular, Laya results retain native SDK truncation.

\begin{table}[t]
\centering
\caption{Binary assessment of local models (\%). Violations are positive and invalid outputs count as errors. Bold marks column maxima. Interface and truncation details are in Appendix~\ref{app:protocol}.}
\label{tab:local_binary}
\footnotesize
\setlength{\tabcolsep}{3pt}
\begin{tabular}{lrrrrrrrr}
\toprule
\rowcolor{TableHeaderGreen}
& \multicolumn{4}{c}{AdaptiveSafety} & \multicolumn{4}{c}{DynaBench} \\
\cmidrule(lr){2-5}\cmidrule(lr){6-9}
\rowcolor{TableHeaderGreen}
Model & Acc. & Prec. & Rec. & F1 & Acc. & Prec. & Rec. & F1 \\
\midrule
Qwen3Guard-Gen-4B & 59.00 & 74.46 & 27.40 & 40.06 & 51.57 & 83.33 & 1.87 & 3.66 \\
\rowcolor{TableBodyGray}
Qwen3Guard-Gen-8B & 58.90 & 75.72 & 26.20 & 38.93 & 51.93 & \textbf{100.00} & 2.25 & 4.40 \\
YuFeng-XGuard-Reason-8B & 57.30 & 66.82 & 29.00 & 40.45 & 50.09 & 48.36 & 22.10 & 30.33 \\
\rowcolor{TableBodyGray}
DynaGuard-4B & 67.30 & 67.76 & 66.00 & 66.87 & 74.22 & 77.25 & 67.42 & 72.00 \\
DynaGuard-8B & 64.10 & 61.97 & 73.00 & 67.03 & \textbf{80.11} & 88.04 & 68.91 & \textbf{77.31} \\
\rowcolor{TableBodyGray}
Laya & 54.40 & 54.98 & 48.60 & 51.59 & 50.46 & 42.86 & 2.25 & 4.27 \\
Laya-Typed-Decisions & 52.30 & 51.26 & \textbf{93.60} & 66.24 & 50.09 & 38.89 & 2.62 & 4.91 \\
\midrule
\rowcolor{TableOursYellow}
AdaGuard-0.6B & 82.60 & 90.55 & 72.80 & 80.71 & 51.38 & 50.37 & \textbf{76.40} & 60.71 \\
\rowcolor{TableOursYellow}
AdaGuard-4B & 89.30 & 95.38 & 82.60 & 88.53 & 71.82 & 70.65 & 73.03 & 71.82 \\
\rowcolor{TableOursYellow}
AdaGuard-8B & \textbf{89.50} & \textbf{96.25} & 82.20 & \textbf{88.67} & 76.80 & 77.22 & 74.91 & 76.05 \\
\bottomrule
\end{tabular}
\end{table}

\begin{figure}[!htbp]
\centering
\includegraphics[width=\textwidth]{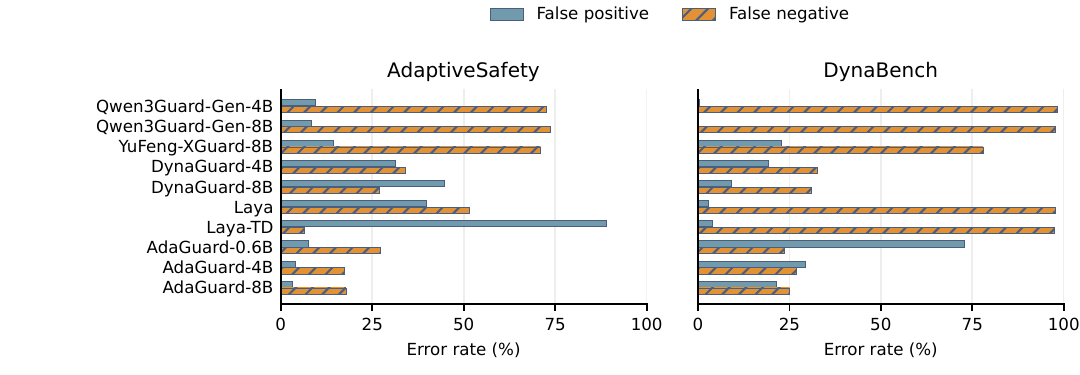}
\caption{False-positive and false-negative rates, normalized by compliant and violating examples, respectively. Invalid outputs count as errors; lower is better.}
\label{fig:error_profiles}
\end{figure}

\paragraph{AdaptiveSafety.}
\sys{}-8B leads the evaluated local models in accuracy, binary F1, exact match, and rule micro-F1 (Tables~\ref{tab:local_binary} and~\ref{tab:local_rules}). Its 89.50\% accuracy and 88.67\% F1 exceed the strongest non-\sys{} baseline on each metric by 22.20 and 21.64 percentage points. The 4B model is close at 89.30\% accuracy and 88.53\% F1. These results extend beyond detecting that some violation occurred. The 8B model identifies the complete rule set in 77.10\% of examples and achieves 74.05\% rule micro-F1. Its 96.25\% binary precision reflects few false alarms, although recall remains lower than Laya-Typed-Decisions. Figure~\ref{fig:error_profiles} shows both error types rather than collapsing this tradeoff into accuracy alone. The advantage on these four endpoints also holds against the API models in Tables~\ref{tab:binary_results} and~\ref{tab:rule_results}; their precision--recall tradeoffs are discussed in Appendix~\ref{app:api}.

\begin{table}[t]
\centering
\caption{Rule identification by local models with policy-local outputs (\%). Exact is complete-set accuracy; Rule-F1 is micro-F1. Binary-only interfaces are excluded. Full rule precision and recall appear in Table~\ref{tab:rule_results}.}
\label{tab:local_rules}
\small
\setlength{\tabcolsep}{3.5pt}
\begin{tabular}{lrrrr}
\toprule
\rowcolor{TableHeaderGreen}
& \multicolumn{2}{c}{AdaptiveSafety} & \multicolumn{2}{c}{DynaBench} \\
\cmidrule(lr){2-3}\cmidrule(lr){4-5}
\rowcolor{TableHeaderGreen}
Model & Exact & Rule-F1 & Exact & Rule-F1 \\
\midrule
Laya & 31.80 & 6.27 & 49.91 & 3.38 \\
\rowcolor{TableBodyGray}
Laya-Typed-Decisions & 8.50 & 8.98 & 49.36 & 2.05 \\
\midrule
\rowcolor{TableOursYellow}
AdaGuard-0.6B & 68.10 & 59.28 & 44.94 & 50.50 \\
\rowcolor{TableOursYellow}
AdaGuard-4B & 76.60 & 71.50 & 63.72 & 54.99 \\
\rowcolor{TableOursYellow}
AdaGuard-8B & \textbf{77.10} & \textbf{74.05} & \textbf{70.72} & \textbf{63.42} \\
\bottomrule
\end{tabular}
\end{table}

\paragraph{Evaluation beyond AdaptiveSafety.}
DynaBench tests compliance with a different collection of user-defined rules (Table~\ref{tab:local_binary}). \sys{}-8B achieves 76.80\% accuracy and 76.05\% F1, substantially above Qwen3Guard-Gen and YuFeng-XGuard under the recorded protocols. It supports explicit rule attribution at 70.72\% exact match, whereas the DynaGuard adapter returns only a binary verdict. DynaGuard-8B remains stronger in accuracy and F1 at 80.11\% and 77.31\%, but has lower recall, 68.91\% versus 74.91\%. Thus, \sys{} supports both detection and rule attribution beyond AdaptiveSafety, with a remaining gap to the specialized DynaGuard model in aggregate binary performance. This evaluation does not establish zero-shot transfer or uniform superiority.

\begin{figure}[!htbp]
\centering
\includegraphics[width=\textwidth]{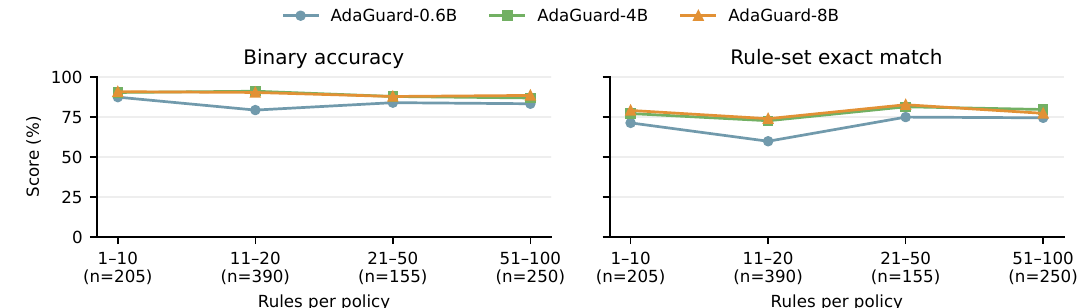}
\caption{AdaptiveSafety by policy length. Counts give group sizes. Lines connect different example groups, not controlled policy changes.}
\label{fig:policy_length}
\end{figure}

\paragraph{Policy length and model size.}
Figure~\ref{fig:policy_length} shows that the 4B and 8B models retain strong AdaptiveSafety performance across policy lengths. With 51--100 rules, the 8B model reaches 88.40\% accuracy and 77.20\% exact match. Exact match is not monotonic in policy length, since the groups differ in content and label composition. The 0.6B model is less consistent, particularly on DynaBench. Appendix~\ref{app:scope} separates user-only requests from trajectories, while Table~\ref{tab:coverage} reports coverage and macro metrics for all models.

\begin{figure}[!htbp]
\centering
\includegraphics[width=\textwidth]{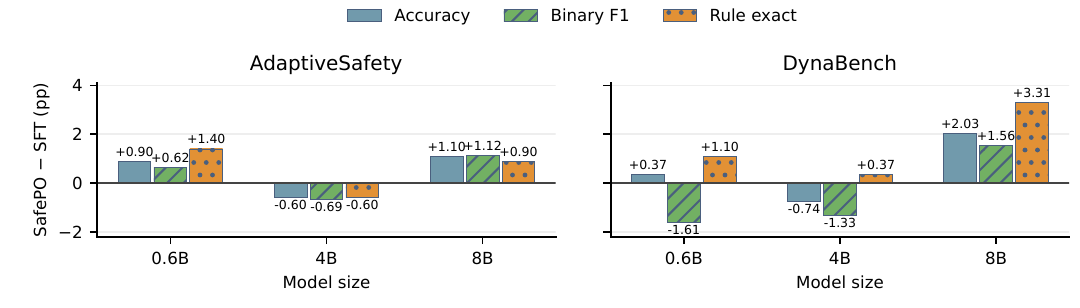}
\caption{SafePO minus historical SFT in percentage points. Positive values indicate improvement. Evaluator differences are detailed in Appendix~\ref{app:sft}.}
\label{fig:training_stages}
\end{figure}

\paragraph{Effect of the training stage.}
Figure~\ref{fig:training_stages} compares the saved SFT and SafePO checkpoints. At 8B, the later checkpoint improves accuracy, binary F1, and rule exact match on both datasets. DynaBench shows the largest exact-match gain, from 67.40\% to 70.72\%, alongside a 2.03-point accuracy gain. Changes at smaller sizes are mixed. At 4B, AdaptiveSafety accuracy decreases by 0.60 points, while DynaBench exact match increases by 0.37 points despite lower binary F1. This supports the 8B checkpoint's improvement across both tasks, but not a universal benefit at every scale. The historical SFT evaluator used a different input-budget calculation; this stage comparison is descriptive, not a controlled component ablation. Appendix~\ref{app:sft} provides the absolute scores and protocol differences.

\section{Related Work}
\label{sec:related_work}

Adaptive guards assess safety under application-specific criteria, with YuFeng-XGuard supporting dynamic policy enforcement and DynaGuard evaluating text against user-defined policies~\citep{lin2026yufengxguard,hoover2026dynaguard}.
Jev and Laya offer a complementary approach through typed decision interfaces, allowing developers to express policy checks as questions with probabilistic outputs~\citep{jevmodel2026,convaiinnovations2026laya}.
These approaches establish flexible policy specification and decision interfaces, while our work focuses on identifying the complete set of rules violated by agent behavior under a supplied policy.
We introduce AdaptiveSafety, which uses policy and behavioral counterfactuals to connect changes in governing rules or agent actions to the corresponding violation sets.
We further develop SafePO, combining structured verdict rewards with value-guided token weighting to train the AdaGuard family for policy-conditioned safety assessment and rule identification.

\section{Conclusion}
\label{sec:conclusion}
We introduced AdaptiveSafety, SafePO, and the \sys{} model family for assessing agent behavior under user-defined policies. The dataset links changes in policy and behavior to complete violation sets, while SafePO combines structured verdict rewards with value-guided weighting of analysis and verdict tokens. \sys{} achieves strong detection and rule-identification results on AdaptiveSafety and supports both tasks on DynaBench. The remaining DynaBench gap, mixed SFT-to-RL changes, and unverified explanation quality delimit these findings. Controlled component ablations and broader deployment studies are needed to establish which training choices drive the gains and how reliably the resulting guards support intervention. Deployment should retain human review for consequential decisions and assess performance under the policies of the target application.

\clearpage
\bibliography{references}
\bibliographystyle{adaguard}

\clearpage
\appendix
\setcounter{table}{0}
\renewcommand{\thetable}{A\arabic{table}}
\renewcommand{\theHtable}{appendix.\arabic{table}}
\setcounter{figure}{0}
\renewcommand{\thefigure}{A\arabic{figure}}
\renewcommand{\theHfigure}{appendix.\arabic{figure}}
\section{Additional Experimental Details}
\label{app:protocol}

\subsection{API-Based Comparisons}
\label{app:api}
Tables~\ref{tab:binary_results} and~\ref{tab:rule_results} extend the binary and rule comparisons in Tables~\ref{tab:local_binary} and~\ref{tab:local_rules} to GPT-5.2, Gemini-3.1-Flash-Lite, and the hosted Jev model. On AdaptiveSafety, \sys{}-8B exceeds all three in accuracy, binary F1, exact match, and rule micro-F1. Relative to Jev, the strongest of these baselines in binary accuracy and F1, the margins are 7.10 and 5.30 percentage points. GPT-5.2 has higher recall at 93.80\%, compared with 82.20\% for \sys{}-8B. On DynaBench, Jev has the highest accuracy at 83.61\%, and GPT-5.2 has the highest F1 at 81.30\%; both exceed \sys{}-8B. These comparisons use the same examples but different model interfaces and inference budgets.

\begin{table}[H]
\centering
\caption{Binary safety assessment (\%). Unsafe is the positive class; invalid outputs count as errors. Bold marks the highest value in each column among the reported runs. Laya results use native SDK truncation.}
\label{tab:binary_results}
\footnotesize
\setlength{\tabcolsep}{2.5pt}
\begin{tabular}{lrrrrrrrr}
\toprule
\rowcolor{TableHeaderGreen}
& \multicolumn{4}{c}{AdaptiveSafety} & \multicolumn{4}{c}{DynaBench} \\
\cmidrule(lr){2-5}\cmidrule(lr){6-9}
\rowcolor{TableHeaderGreen}
Model & Acc. & Prec. & Rec. & F1 & Acc. & Prec. & Rec. & F1 \\
\midrule
GPT-5.2 & 76.10 & 69.28 & \textbf{93.80} & 79.69 & 83.06 & 88.89 & 74.91 & \textbf{81.30} \\
\rowcolor{TableBodyGray}
Gemini-3.1-Flash-Lite & 72.10 & 75.52 & 65.40 & 70.10 & 80.85 & 80.07 & \textbf{81.27} & 80.67 \\
Jev & 82.40 & 79.03 & 88.20 & 83.36 & \textbf{83.61} & 93.20 & 71.91 & 81.18 \\
\rowcolor{TableBodyGray}
Qwen3Guard-Gen-4B & 59.00 & 74.46 & 27.40 & 40.06 & 51.57 & 83.33 & 1.87 & 3.66 \\
Qwen3Guard-Gen-8B & 58.90 & 75.72 & 26.20 & 38.93 & 51.93 & \textbf{100.00} & 2.25 & 4.40 \\
\rowcolor{TableBodyGray}
YuFeng-XGuard-Reason-8B & 57.30 & 66.82 & 29.00 & 40.45 & 50.09 & 48.36 & 22.10 & 30.33 \\
DynaGuard-4B & 67.30 & 67.76 & 66.00 & 66.87 & 74.22 & 77.25 & 67.42 & 72.00 \\
\rowcolor{TableBodyGray}
DynaGuard-8B & 64.10 & 61.97 & 73.00 & 67.03 & 80.11 & 88.04 & 68.91 & 77.31 \\
Laya & 54.40 & 54.98 & 48.60 & 51.59 & 50.46 & 42.86 & 2.25 & 4.27 \\
\rowcolor{TableBodyGray}
Laya-Typed-Decisions & 52.30 & 51.26 & 93.60 & 66.24 & 50.09 & 38.89 & 2.62 & 4.91 \\
\midrule
\rowcolor{TableOursYellow}
AdaGuard-0.6B & 82.60 & 90.55 & 72.80 & 80.71 & 51.38 & 50.37 & 76.40 & 60.71 \\
\rowcolor{TableOursYellow}
AdaGuard-4B & 89.30 & 95.38 & 82.60 & 88.53 & 71.82 & 70.65 & 73.03 & 71.82 \\
\rowcolor{TableOursYellow}
AdaGuard-8B & \textbf{89.50} & \textbf{96.25} & 82.20 & \textbf{88.67} & 76.80 & 77.22 & 74.91 & 76.05 \\
\bottomrule
\end{tabular}
\end{table}

\begin{table}[H]
\centering
\caption{Policy-local rule identification (\%). Exact requires the complete valid violation set; micro scores pool rule decisions within each sample before aggregation. Failed predictions contribute an empty set to micro counts and never count as exact. Binary-only adapters are omitted. Bold marks the column maximum.}
\label{tab:rule_results}
\footnotesize
\setlength{\tabcolsep}{2.5pt}
\begin{tabular}{lrrrrrrrr}
\toprule
\rowcolor{TableHeaderGreen}
& \multicolumn{4}{c}{AdaptiveSafety} & \multicolumn{4}{c}{DynaBench} \\
\cmidrule(lr){2-5}\cmidrule(lr){6-9}
\rowcolor{TableHeaderGreen}
Model & Exact & Micro-P & Micro-R & Micro-F1 & Exact & Micro-P & Micro-R & Micro-F1 \\
\midrule
GPT-5.2 & 52.70 & 44.17 & \textbf{78.18} & 56.45 & \textbf{82.87} & \textbf{88.50} & 74.91 & \textbf{81.14} \\
\rowcolor{TableBodyGray}
Gemini-3.1-Flash-Lite & 61.00 & 64.80 & 42.31 & 51.20 & 80.85 & 80.07 & \textbf{81.27} & 80.67 \\
Jev & 57.60 & 39.70 & 74.82 & 51.88 & 80.85 & 86.82 & 71.54 & 78.44 \\
\rowcolor{TableBodyGray}
Laya & 31.80 & 3.34 & 50.07 & 6.27 & 49.91 & 6.82 & 2.25 & 3.38 \\
Laya-Typed-Decisions & 8.50 & 4.86 & 59.59 & 8.98 & 49.36 & 12.00 & 1.12 & 2.05 \\
\midrule
\rowcolor{TableOursYellow}
AdaGuard-0.6B & 68.10 & 71.73 & 50.51 & 59.28 & 44.94 & 40.78 & 66.29 & 50.50 \\
\rowcolor{TableOursYellow}
AdaGuard-4B & 76.60 & 81.73 & 63.54 & 71.50 & 63.72 & 51.64 & 58.80 & 54.99 \\
\rowcolor{TableOursYellow}
AdaGuard-8B & \textbf{77.10} & \textbf{85.01} & 65.59 & \textbf{74.05} & 70.72 & 61.11 & 65.92 & 63.42 \\
\bottomrule
\end{tabular}
\end{table}

\clearpage
\subsection{Model Interfaces and Inference Settings}
The evaluation files name AdaptiveSafety as \texttt{CUA\_exea\_policy}; all such records refer to the same 1,000-example test set used here. Historical SFT records call this split \texttt{validation}. We use it as the test set; checkpoint selection uses development data drawn separately from the training pool. The policy-free CUA-Exec evaluation is not included in these experiments.

Qwen3Guard-Gen receives the policy as context but retains its fixed safety taxonomy. Its \texttt{Controversial} and \texttt{Unsafe} judgments map to the positive class. YuFeng-XGuard-Reason uses a dynamic category for violation of any supplied rule, with \texttt{a} indicating a violation and \texttt{sec} indicating compliance. DynaGuard retains its native PASS/FAIL interface. These adapters yield binary judgments rather than predicted policy-local rule sets, so Table~\ref{tab:rule_results} does not assign them rule-identification scores.

\sys{}, GPT-5.2, and Gemini use the overall analysis followed by the ordered label sequence defined in Section~\ref{sec:guard_output}. Parsing checks the output tags, policy membership, uniqueness, and rule order. A correct set serialized in the wrong order is invalid at evaluation, although the training reward distinguishes such responses from predictions with incorrect membership. Jev and Laya instead return one \texttt{noul} probability per rule. Probabilities of at least $0.5$ select the corresponding rule; an empty selection is compliant. Their predicted sets are serialized in policy order. We do not infer an overall risk probability from the maximum rule probability.

Local generative models use greedy decoding in BF16 with one model replica per A100-SXM4-80GB GPU across eight GPUs. The recorded prompt budget is 16,000 tokens. Output budgets are 512 tokens for \sys{}, 1,024 for DynaGuard, and 128 for Qwen3Guard-Gen and YuFeng-XGuard-Reason. No input truncation is recorded for these runs. Laya uses SDK version 0.3.5 with native sequence limits of 512 tokens for Laya and 1,024 for Laya-Typed-Decisions; their question-header limits are 192 and 256 tokens, respectively. Truncation affects at least one question in every AdaptiveSafety example for both checkpoints. On DynaBench, it affects 543 of 543 Laya examples and 454 of 543 Laya-Typed-Decisions examples. These results characterize the recorded SDK configuration and should not be interpreted as full-context capability estimates.

The two closed-source generative models use the API identifiers \texttt{gpt-5.2-1211-global} and \texttt{gemini-3.1-flash-lite-preview}. Both use a 2,048-token output budget, with no client-side input cropping or explicit temperature override. Server-side truncation is unknown. Jev resolves to \texttt{typesafe/jev-1.13-20260917}; its internal context handling is also unknown.

\subsection{Coverage and Metric Definitions}
\label{app:coverage}
Table~\ref{tab:coverage} reports invalid counts, detected truncation, and macro metrics. An invalid response on a compliant example contributes a false positive, and an invalid response on a violating example contributes a false negative. This convention penalizes failures without deleting examples; it does not assign a substantive model prediction to an API error. For rule micro metrics, invalid outputs contribute no predicted rules and all reference violations remain false negatives. Rule identifiers are local to each policy, and counts are aggregated only after comparing each prediction with its own reference set. Two valid empty sets receive exact-match credit.

GPT-5.2 has two invalid AdaptiveSafety outputs; Gemini has 103. Gemini's failures are output-validation failures. The local YuFeng run has 139 invalid AdaptiveSafety outputs and 36 on DynaBench. These protocol-dependent failures are included in the main scores, so coverage is relevant when comparing models. 

\begin{table}[H]\centering\footnotesize
\caption{Coverage and macro metrics (\%). Err. is the number of invalid or failed responses. Trunc. is the fraction with detected input truncation; -- means unknown. M-P, M-R and M-F1 denote binary macro averages.}
\label{tab:coverage}
\setlength{\tabcolsep}{2pt}
\begin{tabular}{lrrrrrrrrrr}\toprule
\rowcolor{TableHeaderGreen}
& \multicolumn{5}{c}{AdaptiveSafety} & \multicolumn{5}{c}{DynaBench} \\
\cmidrule(lr){2-6}\cmidrule(lr){7-11}
\rowcolor{TableHeaderGreen}
Model & Err. & Trunc. & M-P & M-R & M-F1 & Err. & Trunc. & M-P & M-R & M-F1 \\\midrule
GPT-5.2 & 2 & -- & 79.84 & 76.10 & 75.33 & 0 & -- & 83.91 & 82.92 & 82.91 \\
\rowcolor{TableBodyGray}
Gemini-3.1-Flash-Lite & 103 & -- & 72.50 & 72.10 & 71.97 & 0 & -- & 80.85 & 80.85 & 80.85 \\
Jev & 0 & -- & 82.84 & 82.40 & 82.34 & 0 & -- & 85.47 & 83.42 & 83.33 \\
\rowcolor{TableBodyGray}
Qwen3Guard-Gen-4B & 0 & 0.00 & 64.99 & 59.00 & 54.45 & 0 & 0.00 & 67.27 & 50.76 & 35.66 \\
Qwen3Guard-Gen-8B & 0 & 0.00 & 65.55 & 58.90 & 53.98 & 0 & 0.00 & 75.70 & 51.12 & 36.15 \\
\rowcolor{TableBodyGray}
YuFeng-XGuard-Reason-8B & 139 & 0.00 & 60.74 & 57.30 & 53.58 & 36 & 0.00 & 49.48 & 49.64 & 45.73 \\
DynaGuard-4B & 20 & 0.00 & 67.31 & 67.30 & 67.29 & 1 & 0.00 & 74.59 & 74.11 & 74.05 \\
\rowcolor{TableBodyGray}
DynaGuard-8B & 3 & 0.00 & 64.56 & 64.10 & 63.81 & 2 & 0.00 & 81.59 & 79.93 & 79.80 \\
Laya & 0 & 100.00 & 54.46 & 54.40 & 54.25 & 0 & 100.00 & 46.76 & 49.67 & 35.43 \\
\rowcolor{TableBodyGray}
Laya-Typed-Decisions & 0 & 100.00 & 57.24 & 52.30 & 42.49 & 0 & 83.61 & 44.68 & 49.32 & 35.54 \\
\rowcolor{TableOursYellow}
AdaGuard-0.6B & 25 & 0.00 & 83.90 & 82.60 & 82.43 & 7 & 0.00 & 52.36 & 51.79 & 48.47 \\
\rowcolor{TableOursYellow}
AdaGuard-4B & 4 & 0.00 & 90.02 & 89.30 & 89.25 & 2 & 0.00 & 71.84 & 71.84 & 71.82 \\
\rowcolor{TableOursYellow}
AdaGuard-8B & 8 & 0.00 & 90.36 & 89.50 & 89.44 & 1 & 0.00 & 76.81 & 76.76 & 76.77 \\
\bottomrule\end{tabular}\end{table}

\subsection{Supervised and Reinforcement-Learned Checkpoints}
\label{app:sft}
Table~\ref{tab:sft_comparison} compares the saved SFT results with the current SafePO checkpoints. The dataset hashes agree across these runs. For 8B, AdaptiveSafety accuracy increases from 88.40\% to 89.50\%, and exact match from 76.20\% to 77.10\%. On DynaBench, accuracy increases from 74.77\% to 76.80\%, and exact match from 67.40\% to 70.72\%.

The changes are not uniformly positive. For 4B, AdaptiveSafety accuracy decreases from 89.90\% to 89.30\% and exact match from 77.20\% to 76.60\%. DynaBench binary F1 decreases from 73.15\% to 71.82\%, although exact match increases from 63.35\% to 63.72\%. For 0.6B, DynaBench F1 decreases from 62.32\% to 60.71\%. These observations do not support a claim that SafePO improves every metric at every model size.

The SFT results are historical evaluations, not checkpoints rerun under the current evaluator. In particular, the earlier AdaptiveSafety input budget included the reference target length, whereas the current evaluator budgets from the input alone. The historical AdaptiveSafety summaries record zero truncation, but the preprocessing implementations still differ. We therefore treat this comparison as descriptive evidence across saved checkpoints, not a controlled estimate of the effect of SafePO. No component ablations isolating the reward terms, value modulation, or region budgets are available in the supplied results.

\begin{table}[H]\centering\small
\caption{Historical SFT and current SafePO checkpoints (\%). These are descriptive comparisons, not controlled component ablations. The input files match, but the historical preprocessing budget depended on the target length.}
\label{tab:sft_comparison}
\begin{tabular}{llrrrrrr}\toprule
\rowcolor{TableHeaderGreen}
& & \multicolumn{3}{c}{AdaptiveSafety} & \multicolumn{3}{c}{DynaBench} \\
\cmidrule(lr){3-5}\cmidrule(lr){6-8}
\rowcolor{TableHeaderGreen}
Size & Stage & Acc. & F1 & Exact & Acc. & F1 & Exact \\\midrule
\rowcolor{TableOursYellow}
0.6B & SFT & 81.70 & 80.09 & 66.70 & 51.01 & 62.32 & 43.83 \\
\rowcolor{TableOursYellow}
0.6B & SafePO & 82.60 & 80.71 & 68.10 & 51.38 & 60.71 & 44.94 \\
\rowcolor{TableOursYellow}
4B & SFT & 89.90 & 89.22 & 77.20 & 72.56 & 73.15 & 63.35 \\
\rowcolor{TableOursYellow}
4B & SafePO & 89.30 & 88.53 & 76.60 & 71.82 & 71.82 & 63.72 \\
\rowcolor{TableOursYellow}
8B & SFT & 88.40 & 87.55 & 76.20 & 74.77 & 74.49 & 67.40 \\
\rowcolor{TableOursYellow}
8B & SafePO & 89.50 & 88.67 & 77.10 & 76.80 & 76.05 & 70.72 \\
\bottomrule\end{tabular}\end{table}

\subsection{Assessment Scope}
\label{app:scope}
Table~\ref{tab:scope} separates the 117 user-only requests from the 883 trajectories in AdaptiveSafety. \sys{}-4B reaches 91.45\% accuracy on requests and 89.01\% on trajectories; \sys{}-8B reaches 90.60\% and 89.35\%, respectively. The 4B model has higher request-level exact match, while 8B has higher trajectory-level exact match. The combined result therefore covers two assessment scopes rather than only agent executions.

\begin{table}[H]\centering\footnotesize
\caption{AdaptiveSafety by assessment scope (\%). Query-only inputs assess the request; trajectory inputs assess recorded agent behavior.}
\label{tab:scope}
\begin{tabular}{lrrrrrr}\toprule
\rowcolor{TableHeaderGreen}
& \multicolumn{3}{c}{Query (117)} & \multicolumn{3}{c}{Trajectory (883)} \\
\cmidrule(lr){2-4}\cmidrule(lr){5-7}
\rowcolor{TableHeaderGreen}
Model & Acc. & F1 & Exact & Acc. & F1 & Exact \\\midrule
GPT-5.2 & 81.20 & 75.56 & 71.79 & 75.42 & 80.04 & 50.17 \\
\rowcolor{TableBodyGray}
Gemini-3.1-Flash-Lite & 70.09 & 60.67 & 61.54 & 72.37 & 71.09 & 60.93 \\
Jev & 82.91 & 71.43 & 70.94 & 82.33 & 84.21 & 55.83 \\
\rowcolor{TableBodyGray}
Laya & 65.81 & 39.39 & 56.41 & 52.89 & 52.51 & 28.54 \\
Laya-Typed-Decisions & 41.03 & 48.89 & 15.38 & 53.79 & 68.08 & 7.59 \\
\rowcolor{TableOursYellow}
AdaGuard-0.6B & 83.76 & 71.64 & 74.36 & 82.45 & 81.44 & 67.27 \\
\rowcolor{TableOursYellow}
AdaGuard-4B & 91.45 & 86.11 & 85.47 & 89.01 & 88.73 & 75.42 \\
\rowcolor{TableOursYellow}
AdaGuard-8B & 90.60 & 84.51 & 81.20 & 89.35 & 89.02 & 76.56 \\
\bottomrule\end{tabular}\end{table}

\subsection{Interpretation and Remaining Evidence}
All reported evaluations are single runs. We report descriptive differences without significance claims or estimates of training-seed variability. The analysis measures verdict accuracy and rule identification; it does not independently verify the factual correctness of generated explanations. Policy-length groups differ in content and label composition, and the comparisons across model families use different interfaces and context limits. Controlled component ablations and matched-protocol SFT comparisons remain necessary to attribute gains to individual SafePO mechanisms.

\subsection{Reproducibility Resources}
\label{app:resources}
The project repository is available at {\hypersetup{colorlinks=true,urlcolor=blue}\url{https://github.com/Yunhao-Feng/AdaGuard}}. Section~\ref{sec:data_and_initialization} describes dataset construction and supervised initialization, and Sections~\ref{sec:safepo} and~\ref{sec:region_value_weighting} specify SafePO. Appendix~\ref{app:protocol} records the evaluation interfaces and inference settings used for the reported results.

\section{Value Learning and Modulation Details}
\label{app:value_details}
We use the bounded prediction
$V_{i,t}=(2/\pi)\arctan z_\phi(x_i,y_{i,<t})$
and minimize
\begin{equation}
    \mathcal{L}_V
    =
    \frac{1}{2B}
    \sum_{i,t}w^0_{i,t}
    \max\left\{
        (V_{i,t}-R_i)^2,
        (\widetilde{V}_{i,t}-R_i)^2
    \right\},
    \label{eq:safepo_value_loss}
\end{equation}
where
$\widetilde{V}_{i,t}
=\operatorname{clip}(V_{i,t},
\overline{V}_{i,t}-0.2,\overline{V}_{i,t}+0.2)$
and $\overline{V}_{i,t}$ denotes the detached pre-update prediction.
The value sequence shown as $v_i$ in Figure~\ref{fig:adaguard_framework} is $v_i=(\overline{V}_{i,1},\ldots,\overline{V}_{i,T_i})$; its adjacent differences provide the $\Delta V$ signal used for token weighting.
Every prefix uses the final response reward as its target.
The value model has its own parameters and leaves the response-level advantage unchanged.

We control the modulation strength using the value model's predictive performance.
Let $E_V$ be its pre-update mean squared error aggregated with $w^0$, and let $E_G$ be the mean squared error of predicting each response reward from the mean reward of the other responses in its group.
With exponential moving averages of decay $0.95$, we set
\begin{equation}
    \kappa_{\mathrm{next}}
    =
    \operatorname{clip}
    \left(
        1-
        \frac{\operatorname{EMA}(E_V)}
             {\operatorname{EMA}(E_G)+10^{-8}},
        0,1
    \right).
    \label{eq:safepo_modulation_control}
\end{equation}
We initialize $\kappa=0$ and disable modulation when the group-error estimate is numerically negligible.
The updated coefficient applies only to the next batch.
Thus, an insufficiently predictive value model yields a fallback to fixed region weighting.

\section{Supervised Training Objective}
\label{app:sft_objective}
Full-parameter supervised training minimizes
\begin{equation}
    \mathcal{L}_{\mathrm{SFT}}
    =
    -
    \frac{
        \sum_{n,t}\lambda_{n,t}
        \log\pi_\theta
        (y^\star_{n,t}\mid x_n,y^\star_{n,<t})
    }{
        \sum_{n,t}\lambda_{n,t}
    },
    \label{eq:weighted_sft}
\end{equation}
where the sums range over a minibatch.
The weights are $4$ on the complete label span, $1$ on other response tokens, and $0$ on prompt and padding positions.

\end{document}